# ROBUST REAL TIME FACE RECOGNITION AND TRACKING ON GPU USING FUSION OF RGB AND DEPTH IMAGE.

Narmada Naik[1] and Dr.G.N Rathna[2]

[1]Department of Electrical Engineering, Indian Institute of science, Bangalore, India
nnreema22@gmail.com
[2]Department of Electrical Engineering, Indian Institute of science, Bangalore, India
rathna@ee.iisc.ernet.in

*Abstract*

*This paper presents a real-time face recognition system using kinect sensor. The algorithm is implemented on GPU using opencl and significant speed improvements are observed. We use kinect depth image to increase the robustness and reduce computational cost of conventional LBP based face recognition. The main objective of this paper was to perform robust, high speed fusion based face recognition and tracking.*

*The algorithm is mainly composed of three steps. First step is to detect all faces in the video using viola jones algorithm. The second step is online database generation using a tracking window on the face. A modified LBP feature vector is calculated using fusion information from depth and greyscale image on gpu. This feature vector is used to train a svm classifier. Third step involves recognition of multiple faces based on our modified feature vector.*

**KEYWORDS**

*GPU, OpenCL, Face detection, Tracking, LBP, Histogram, SVM, Depth Camera (Kinect).*

## 1. INTRODUCTION

Face recognition is the challenging field in pattern recognition. Face recognition appears to offer more advantages than other biometrics [1-2]. Each biometric has its own merits and demerits. Face recognition can be done implicitly i.e. the face image can be captured from certain distance with less expensive equipment's, whereas iris recognition can't be done implicitly and requires expensive equipment's for recognition. Similarly, for speech recognition which is susceptible to noise, creating distortion on the original speech, whereas face recognition is less susceptible to noise. In the past, couple of decades various methods of face recognition has been proposed and survey can be found in [3-6]. With the development of stochastic machine learning, there are lots of face recognition algorithms based on local features [7].

After being integrated with mobile services and social networking for real time application face recognition has received significant attention in recent years [8]. Despite progress in face recognition field, exact face recognition remains a challenging task in real time environment. In videos the face expression, lighting conditions and location varies from frame to frame [9-10]. The facial expression and pose are the most important aspect for video based face recognition systems, with the change in expression or pose the accuracy of the recognition system decreases, thereby giving false recognition. Therefore, in this paper we have done face recognition based on fusion of RGB and depth image. Fusion based region of interest removes undesired effects due to pose variations and illumination changes [11].

A typical video based face recognition system consist of face detection, face tracking, feature extraction, training and classification. Here, each individual need to be tracked from different view angle and need to be processed at the current frame rate. This real time face recognition system thereby process huge amount of data and requires to be fast. Since, face recognition algorithm are feasible for parallel processing. Therefore, in this paper the implementation was done on GPU which is a parallel device using heterogeneous computing language opencl.

In this paper, we present online or real time video based face recognition of multiple faces, using Kinect sensor. In the detection stage the face was detected on RGB camera, and affine transformation was done to get the depth image of the RGB detected face. The rest of the paper is arranged as follows: Section 2 Overview of the algorithm, where it is discussed briefly about detection of face using depth camera, face tracking, feature extraction and SVM classification for multiple face recognition. Section 3 Implementation GPU-CPU based, Section 4 Experimental results. Section 5 Conclusion.

## OVERVIEW OF ALGORITHM

### 2.1 Face detection and RGB to depth transformation.

Face detection is not straightforward because it has lot of variation, such as pose variation, image orientation, illumination condition and facial expression [12]. Many novel approaches have been already proposed. In this paper, face detection with RGB image was done using viola jones algorithm [13], then the corresponding depth image of the detected face in RGB was calculated using camera transformation and affine transformation. Camera transformation was implemented by applying one to one mapping from RGB detected face enclosed in a rectangle to its depth coordinate system by using Kinect sensor. The result obtained with one to one mapping gives us a rhombus enclosed detected face because of camera transformation [14]. Since we need a rectangle region of the detected face for LBP feature extraction, we performed affine transformation.

Affine transformation is defined as a function between affine spaces which keeps points, straight lines and planes [15]. It is defined as,

p'=x.u+y.v+t  (1)

Where u and v are basis vectors and t is point, and u, v and t (basis and origin) forms a frame for an affine space, p' represent the affine transformation.

**Pseudocode for RGB to Depth transformation.**

**For i< srcDepth.rows and**

**For j< srcDepth.cols do**

**depth=srcCopyD(i,j)**

**NuiImageGetColorPixelCoordinatesFromDepthPixelAtResolution()**

**if(x>0 && y>0 && x<grayImage.cols && y<grayImage.rows) then**

**if grayImage.(x,y)==255 then**

**grayImageDepth.(x,y)=255**

**push to a vector point**

**findContours();**

**drawContours();**

**fitEllipse(contour);**

**Rect r= boundingRect (Mat (contour));**

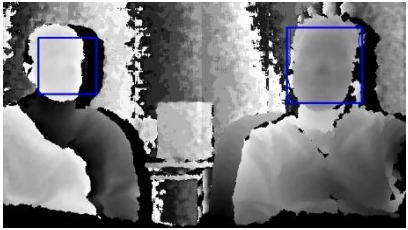 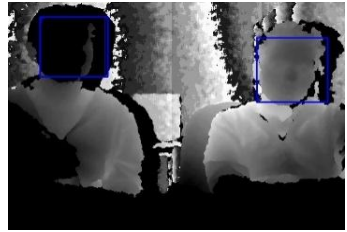

**Fig.1(a)**                         **Fig.1(b)**

**Fig.1(a) is depth image of corresponding RGB image and its face detected region, when object was far away from camera and Fig.1(b) when object was near to camera.**

### 2.2 Face Labelling

Face labelling gives the proper identity of each individual in recognition. In this paper, each detected face is given a label. So, in recognition part the face along with its label get displayed, which will help to distinguish between faces. Since, this paper presents multiple face recognition. Here, each individual face is given a particular label as shown in the Fig.2 and Fig.3.

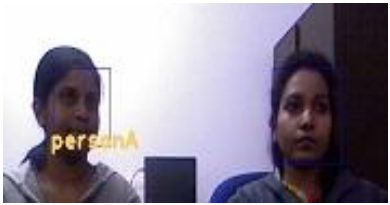 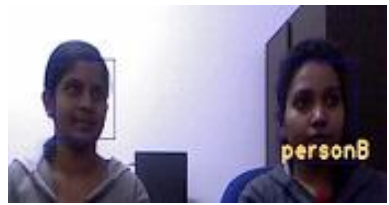

**Fig.2(a) Label for personA.**          **Fig.2(b) Label for personB.**

### 2.3 Face Tracking

The detected face is tracked in order to get multiple data set for single face which increases the accuracy of online face recognition. Person specific face tracking is a challenging task due to changes in environment, such as illumination, and also due to occlusions. Person with specific label or identity is to be tracked for a certain time period [9]. In this paper, tracking algorithm was based on nearest centre of the rectangle which encloses the face in the next frame. If c1 is the centre of rectangle as shown in Fig.4 where the detected face is enclosed, and c2 is the rectangle with detected face in next frame of same personA, whereas c3 and c4 are rectangles of the detected faces of person and personC, then the face of personA is tracked by calculating the nearest center among c2, c3 and c4 of latest frame to c1 in the previous frame. Thereby, all the detected faces are tracked. The data set at each frame gets updated during tracking.

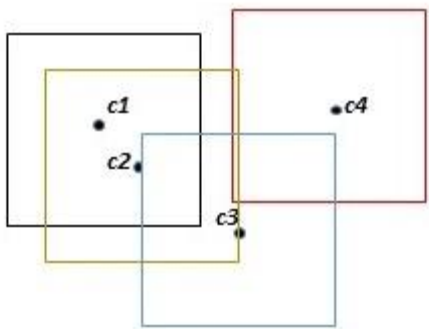

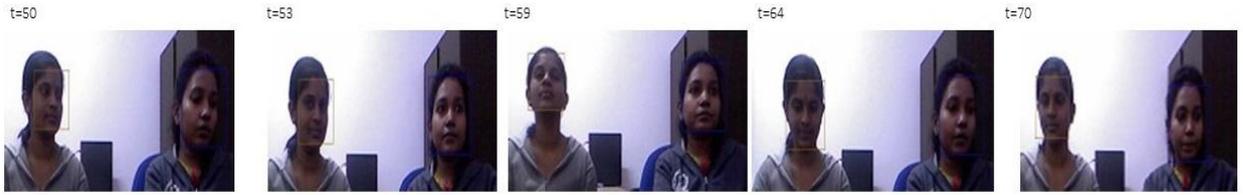

Face tracking for personA

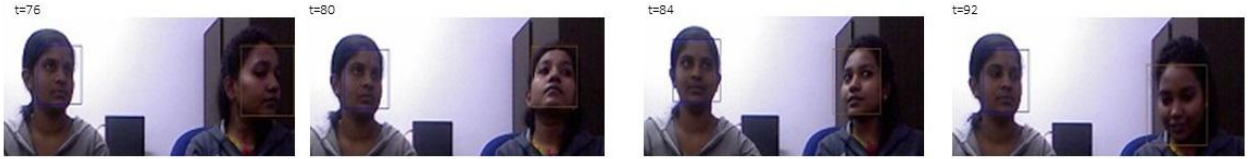

Face tracking for personB.

### 2.4 Feature extraction

Feature extraction is a form of reducing the dimension. Many holistic face descriptor are present such as, PCA(Principal Component Analysis), LDA(Linear Discrimant Analysis)[15], etc. In this paper LBP operator is used for feature extraction. The LBP operator was originally proposed by ojala[16]. LBP operator is one of the best texture descriptors. Its key advantage is, its invariant to monotonic grey level changes. The LBP operator initially threshold the 3x3 neighbourhood pixel based on the centre pixel, and get the decimal value as shown in Fig.8.

The LBP operator is defined as : $LBP_{P,R}(X_C, Y_C) = \sum_{p=o}^{P-1} S(g_p - g_c) 2^P$ (2)

Is the intensity of the image at the Pth sample point where P is the total number of the sample point at a radius of R denoted by(P,R)[17]. The P spaced sampling points of the window are used to calculate the difference between centre gth and its surrounding pixel. Histograms reduces the feature vector length but still retaining 90% of data[17]. Histogram calculation method is given by:

Histogram calculation method is given by: $H_s(p,k) = \sum_{i=1}^{I} \sum_{j=1}^{J} f(LBP(P,R), P)$ (3)

where k is an integer to represent the sub-histograms i.e. obtained from each image k=1,2...K. K is the total no of histograms, and $f(x,y) = \begin{cases} 1, x=y \\ 0, otherwise \end{cases}$ where f(x, y) is the LBP calculated value at pixel(x, y).

![LBP Calculation example]

Pattern = 11110001    LBP = 1 + 16 +32 + 64 + 128 = 241

**Fig.7 LBP Calculation**

## 2.5 Training and classification

With the increase of face recognition application in real world, the accuracy of recognition and training plays a crucial role. Facial expression causes much difficulty in recognition. Therefore, in this paper we considered the depth image, where expressions of the faces can't be seen in depth images as shown in Fig.2 (a). Here, each detected face after the feature has been extracted was trained using support vector machine [18]. Support vector machine is a classifier defined by a hyperplane. SVM is basically finding a hyperplane that gives the largest minimum distance to the training dataset.

In this paper while training personA was considered as class faces and personB face was considered as class nonfaces. Similarly, while training for personB it was considered as class faces and personA as class nonfaces. Classification result is shown in Fig.9.

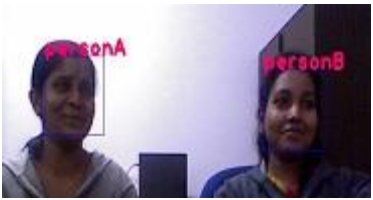

**Fig.9 Multiple recognized faces**.

## 3. IMPLEMENTATION

The main goal of this paper was online multiple face recognition using both color and depth image information from Kinect sensor(Kinect camera). The implementation methodology outlines key activities are: Face detection, Labelling and Tracking , Rgb to depth transformation, Training, Classification. The complete system of face recognition is shown in Fig.9. Here, implementation of the algorithm is done using both OpenCL and OpenCV. LBP feature extraction is computationally expensive, OpenCL is used to accelerate the speed of the feature extraction for multiple data sets.

Here, the face recognition system uses both color and depth image information for detection, tracking, feature extraction. The detected face in color space is resized to 200x200 pixels and is mapped to its depth coordinate system. As, a result we detect the faces in the depth image as shown in Fig.1(a) and Fig.1(b). The detected face is labelled and tracked. Similarly, labelling and tracking for other faces are also done and simultaneously the database gets updated every time a new face is labelled.

After tracking and updating the database, the LBP feature extraction of the detected face and histogram was calculated using OpenCL in GPU. The depth image is ushort type. After LBP feature extraction histogram is calculated ranging from 0 to 255, for each face in GPU. Now, the histogram were used to for SVM training. Then it was tested, both training and testing was done on CPU

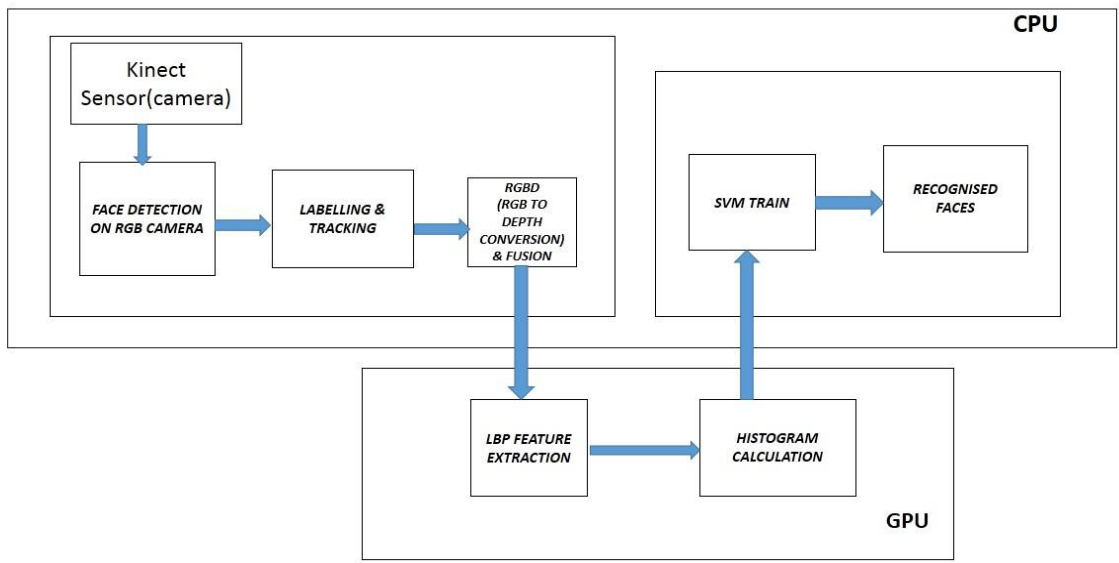

**Fig.10 Block diagram of the Implementation of algorithm discussed**.

## 4. EXPERIMENTAL RESULTS

Results of the experiments with tracking and without tracking using RGB only, and using RGB and depth are shown in Table.1.

| input for feature extraction | Percentage of accuracy (with tracking) | Percentage of accuracy (without tracking) |
|---|---|---|
| Depth Image | 94-98% | 50-60% |
| RGB Image | 70-75% | 40-45% |

**Table.1 Comparison Table**

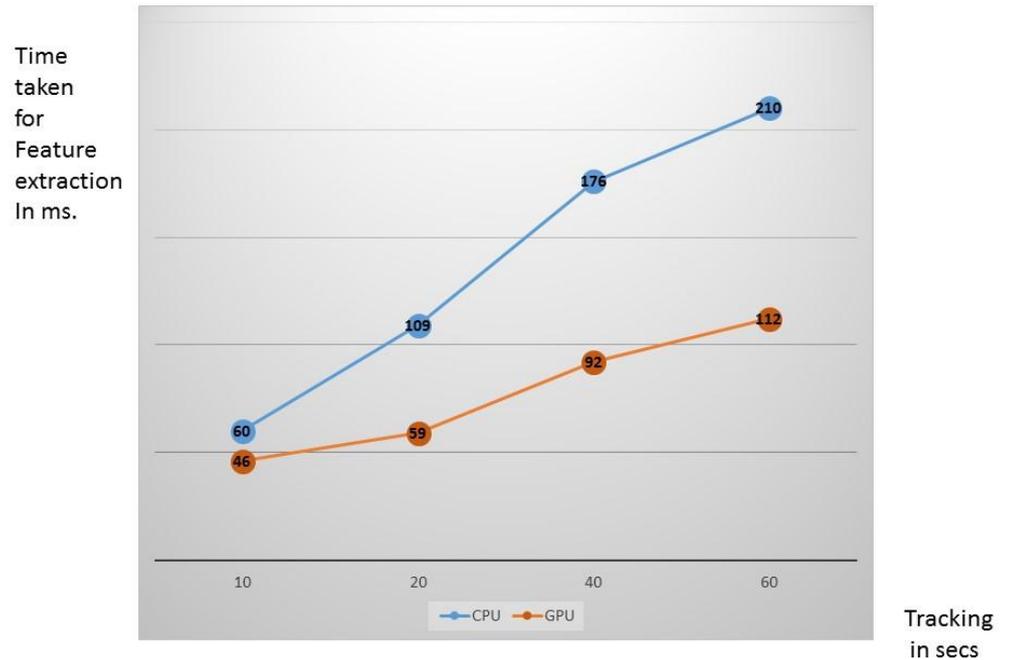

**Fig.11 Performance chart for GPU and CPU for feature extraction, with tracking in secs**.

In this paper, LBP feature extraction was implemented both on cpu and gpu to compare the performance. Experiment Results are shown in Fig.11.

## 5. CONCLUSIONS

This paper discuss implementation of fusion based robust real time multiple face recognition tracking and LBP feature extraction on gpu. All of the modules are the integral and important part in this algorithm for multi face recognition, because accurate face classification depends on this. The modules shown in Fig.10 was implemented on AMD 7670M GPU, i5 3rd generation CPU. Experiments have been performed and results indicate that our system is efficient and accurate for realtime multiple face recognition.

**Authors**

Short Biography

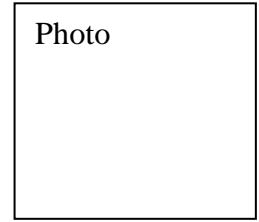

Photo